\documentclass[english]{IEEEtran}
\usepackage[T1]{fontenc}
\usepackage[latin9]{inputenc}
\usepackage{color}
\usepackage{array}
\usepackage{url}
\usepackage{amsmath}
\usepackage{graphicx}
\PassOptionsToPackage{normalem}{ulem}
\usepackage{ulem}

\makeatletter

\providecommand{\tabularnewline}{\\}

\numberwithin{equation}{section}
\numberwithin{figure}{section}

\usepackage{cite}
\counterwithout{figure}{section}
\counterwithout{table}{section}

\makeatother

\usepackage{babel}
\begin{document}
\title{The TerraByte Client: providing access to terabytes of plant data}

\author{\IEEEauthorblockN{Michael A. Beck\IEEEauthorrefmark{1}, Christopher P. Bidinosti, Christopher J. Henry, Manisha
Ajmani\\}
\IEEEauthorblockA{University of Winnipeg, Winnipeg, MB Canada\\
Email: \IEEEauthorrefmark{1}m.beck@uwinnipeg.ca}}
\maketitle
\begin{abstract}
In this paper we demonstrate the TerraByte Client, a software to download
user-defined plant datasets from a data portal hosted at Compute Canada.
To that end the client offers two key functionalities: (1) It allows
the user to get an overview on what data is available and a quick
way to visually check samples of that data. For this the client receives
the results of queries to a database and displays the number of images
that fulfill the search criteria. Furthermore, a sample can be downloaded
within seconds to confirm that the data suits the user's needs. (2)
The user can then download the specified data to their own drive.
This data is prepared into chunks server-side and sent to the user's
end-system, where it is automatically extracted into individual files.
The first chunks of data are available for inspection after a brief
waiting period of a minute or less depending on available bandwidth
and type of data. The TerraByte Client has a full graphical user interface
for easy usage and uses end-to-end encryption. The user interface
is built on top of a low-level client. This architecture in combination
of offering the client program open-source makes it possible for the
user to develop their own user interface or use the client's functionality
directly. An example for direct usage could be to download specific
data on demand within a larger application, such as training machine
learning models. 
\end{abstract}

\section*{Availability}

The \emph{TerraByte Client }is available at the TerraByte homepage
\textcolor{blue}{\uline{\url{https://terrabyte.acs.uwinnipeg.ca/resources.html}}}. 

\section{Introduction\label{sec:Introduction}}

Digital agriculture is an upcoming field with the goal of using modern
technologies such as robotics, phenotyping, and machine learning to
advance and transform the way we grow food \cite{doi:10.5772/62059,BECHAR2017110,BECHAR2017110part2,20193426318,DBLP:journals/corr/abs-1806-06762,RELFECKSTEIN2019100307,BACCO2019100009}.
The vision of autonomous agents making plant-by-plant decisions within
the field motivates this research. However, we often encounter, especially
in the area of machine learning, a bottleneck of \emph{labelled data}
on which models can be trained and improved upon \cite{s18082674,JHA20191}.
Labelled data, at a minimum, is the data itself (for example an image)
plus information about what the data represents (for example the label
``Soybean'' attached to an image). Often, the information provided
-- metadata -- goes beyond simple labelling (for example also tracking
the age of the Soybean plant, time and day of imaging, geolocation,
etc.). In machine learning, for example with convolutional neural
networks (CNN), we require thousands if not hundreds of thousands
of labelled images to perform well. Labelled data is often created
by manually labelling existing datasets and scaled to millions of
images by crowdsourcing the efforts \cite{russell2008labelme,buhrmester2016amazon,rapson2018reducing}.
However, in the case of plant identification, the knowledge required
for accurate labelling is more complex and uncommon and thus this
approach of crowdsourcing is not available. 

Still, when looking at similar applications we can expect strong results,
once the challenge of having labelled image-data is resolved. CNNs
have been applied very successfully in image-classification tasks
with results that eventually outperformed human-level accuracy (see
for example \cite{Simonyan2014,Taigman_2014_CVPR,Henry2019}). What
these models have in common is that they were trained upon a vast
amount of image data. The most prominent of these datasets being ImageNet
\cite{He_2015_ICCV,ILSVRC15}, which consists of more than 1.2 million
labelled images. 

\begin{figure}
\centering{}\includegraphics[width=0.95\columnwidth]{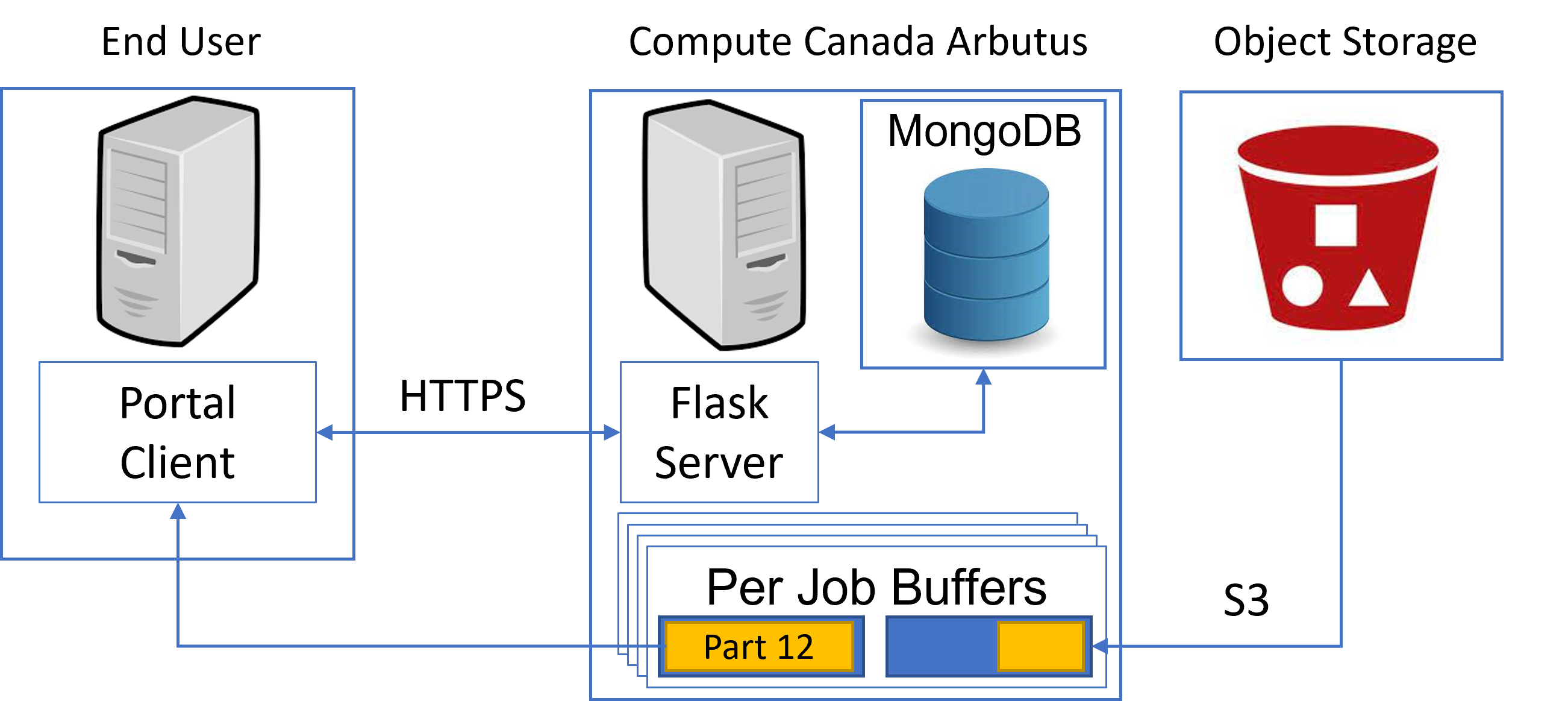}\caption{The architecture that provides plant-data to researchers, consisting
of the TerraByte Client, the server back-end, and an S3-compatible
object storage. In this image the buffer containing part 12 of a dataset
is being downloaded by the client, while in parallel the next part
of the dataset is being fetched from the object storage by the server
into the second buffer.\label{fig:Architecture}}
\end{figure}

At the TerraByte project \cite{TerraByte:Homepage} we have the goal
to provide labelled plant-data at a similar scale, thus, fuelling
the development of better models. For this purpose, we are deploying
and developing several systems to collect plant-data in the lab or
in the field. Since 2019 we have collected over half a million images
in the field and over half a million images in the lab (as of the
end of 2021) \cite{10.1371/journal.pone.0243923}. The images collected
in the lab further contain multiple plants per image, which had been
automatically cropped out, resulting in more than 1.5 million images
of individual plants. All lab-images are fully labelled, providing
information about each plant's species, planting date, imaging date
and more. This data is not only critical for machine learning purposes,
but also allows to filter the data by these metadata fields. The species
of plants, the age of the plant, and the imaging date, are just three
example fields we want to mention here; a full list of the metadata
collected is given in Table \ref{tab:metadata}.

Generating and labelling data alone would not be sufficient, however,
to sustainably support machine learning research in digital agriculture.
As outlined, for example, by Lobet \cite{LOBET2017559}, data-collections
must have long-term access, storage, and curation. To this end we
work together with \emph{Compute Canada}\footnote{Compute Canada's services are actually merged into those of the\emph{
Digital Research Alliance of Canada}. As the day of writing, this
merge has not happened, yet, thus we stick to \emph{Compute Canada}
in this paper. The merger has no effect on where or how to access
our data.} to host our data in a permanent and unchanging location.

To provide this data to other researchers we are performing several
steps: (1) our data is regularly backed up on an object storage system
hosted by \emph{Compute Canada }(CC). (2) The respective metadata
is ingested on a database that runs on a server-instance on CC. This
instance also hosts the back-end application through which data is
delivered from the object storage to the end-user. (3) We have developed
a client program, the \emph{TerraByte Client} (TC), that has been
successfully tested on Windows, MacOS, and Unix. This client communicates
with the back-end application and can download data to the researcher's
system. The TC can also be used to query the database without having
to fully download the dataset in its entirety. Thus, the end-user
can quickly check how many images are available under a specified
set of filters (for example, how many images of 2-3 week old soybeans
are available). Furthermore, samples of such specified datasets can
be downloaded within seconds to confirm that these are indeed images
that suit the end-user's goals. By providing these functions, we think
a dataset of this size is easier to navigate and data relevant to
a specific research question can be found faster. To reduce friction
further, we developed a graphical user interface that enables all
functionality of the TC; we also deliver the program as an installation-free
standalone executable for windows-based machines. 

The rest of this paper is structured as follows: In Section \ref{sec:Operation}
we give an overview on how the TC is operated to obtain data. It follows
a short discussion on the client's performance in Section \ref{sec:Performance}.
The architecture of the TC and the server's API is provided in Section
\ref{sec:Architecture}. Section \ref{sec:Conclusion-and-Future}
offers an outlook on how the TC and the available datasets will evolve. 

\section{Operation\label{sec:Operation}}

\begin{table}
\caption{\label{tab:metadata}}

\begin{centering}
\begin{tabular}{|>{\centering}p{2cm}|>{\centering}p{5cm}|}
\hline 
Field & Description\tabularnewline
\hline 
\hline 
camera \& lens & Name of camera used and lens/model used\tabularnewline
\hline 
camera\_pose & Coordinates and orientation (pan and tilt) of the camera, when the
image was taken\tabularnewline
\hline 
\textbf{date \& time} & \textbf{Date and time when the images was taken}\tabularnewline
\hline 
institute \& room & Location of the imaging system (more than one imaging system will
be in place)\tabularnewline
\hline 
\textbf{horizontal and vertical dimensions} & \textbf{Dimensions of the image taken}\tabularnewline
\hline 
tags & For future use\tabularnewline
\hline 
\end{tabular} \smallskip{}
\par\end{centering}
\begin{centering}
Metadata fields specific for original (uncropped) images.
\par\end{centering}
\bigskip{}

\begin{centering}
\begin{tabular}{|>{\centering}p{2cm}|>{\centering}p{5cm}|}
\hline 
Field & Description\tabularnewline
\hline 
\hline 
\textbf{plant\_id} & \textbf{Unique identifier for each individual plant}\tabularnewline
\hline 
\textbf{label} & \textbf{Common name label for species of plant, e.g. ``Soybean''}\tabularnewline
\hline 
scientific\_name & E.g. ``Fallopia convolvulus''\tabularnewline
\hline 
planting\_date & Date at which the individual plant was planted\tabularnewline
\hline 
\textbf{age} & \textbf{Age of the plant when image was taken in days}\tabularnewline
\hline 
position\_id & An identifier that corresponds to a specific location inside the imagable
volume at which the plant was located when the image was taken\tabularnewline
\hline 
\textbf{x\_min, x\_max, y\_min, y\_max} & \textbf{Coordinates of the subimage within the original image}\tabularnewline
\hline 
\end{tabular}\smallskip{}
\par\end{centering}
\begin{centering}
Metadata fields specfic for cropped out single-plant images
\par\end{centering}
\centering{}(see Figure \ref{fig:An-original-image}).
\end{table}

\begin{figure}
\centering{}\includegraphics[width=0.95\columnwidth]{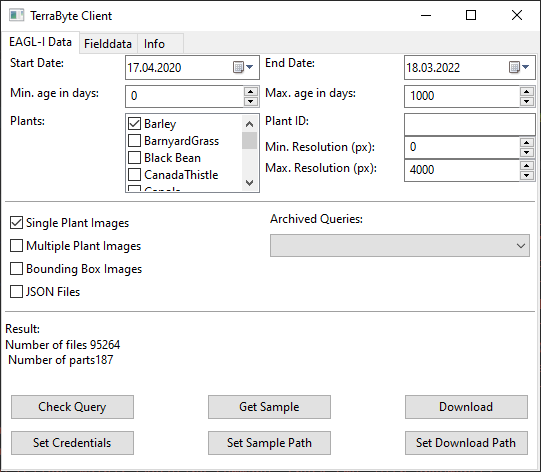}\caption{The TerraByte Client user interface\label{fig:The-TerraByte-Client}
currently showing the \emph{EAGL-I Data }tab, relating to images taken
in the lab.}
\end{figure}

The interface of the TC is shown in Figure \ref{fig:The-TerraByte-Client};
it consists of several tabs each relating to one class of data and
one additional tab providing version-info and useful links. In Figure
\ref{fig:The-TerraByte-Client} the active tab is set to \emph{EAGL-I
Data}, which corresponds to images that have been taken in the lab
by our robotic imager EAGL-I (see \cite{10.1371/journal.pone.0243923}).
The tab itself is structured into three sections; via the top two
sections the user can define the data they are interested in, whereas
the third section contains controls for configuring the tool itself
or for interacting with the server-backend. 

The filtering options of the first section relate directly to the
metadata information that is saved along the image data. Table \ref{tab:metadata}
gives a detailed description of all metadata tracked and we have highlighted
fields in boldface that are currently being used by the TC. All filters
default to the least restrictive setting and combine in an AND-relation.
For example, setting a maximal age of 10 days and a specific plant
ID will only give images of that plant ID in which the plant was less
than 10 days old. Or stated differently, activating additional filters
will generally result in less data fitting the filters. The only exception
to this is the multi-select box in which different plant-species can
be chosen. Choosing no plant at all in this box deactivates the filter
completely, whereas multiple selections will result in every plant
that fits into any of the selected species being returned. Thus, within
the species-filter we have an OR-relation, while the species-filter
itself is combined via AND with the other filters. 

\begin{figure}
\centering{}\includegraphics[width=0.95\columnwidth]{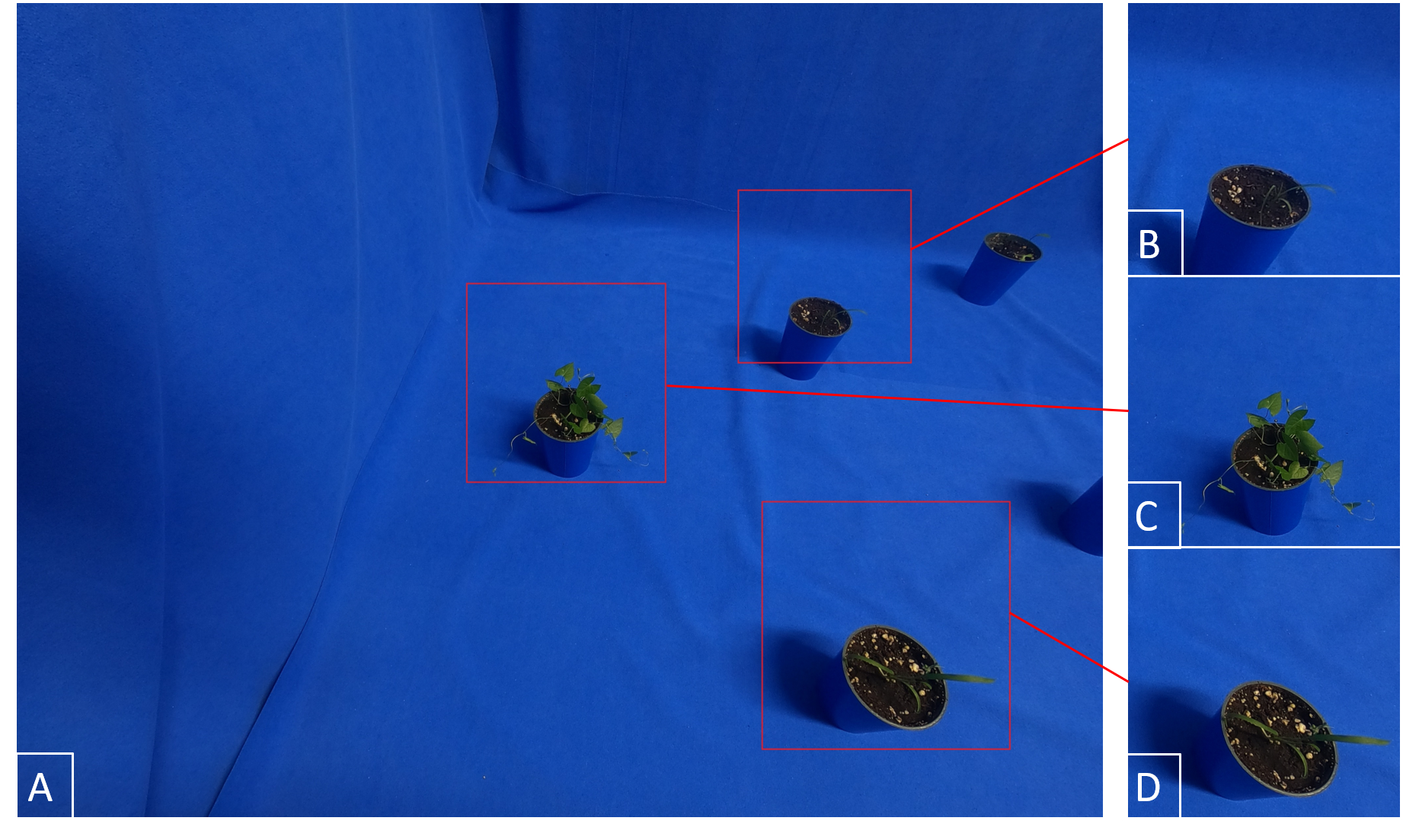}\caption{An original image (A) overlayed with the areas that are being cropped
out as single-plant images (B-D)\label{fig:An-original-image}}
\end{figure}

Through the second section in the interface the user can define what
types of data they are interested in. This relates to the way the
robotic system EAGL-I takes images and we give a brief explanation
of that here: EAGL-I moves a camera inside an imageable volume to
different positions and from each position one image is taken. Each
of those images can contain multiple plants. In a post-processing
step each plant is cropped out from the original image separately.
This usually leads to more than one cropped out plant image per original
image (see Figure \ref{fig:An-original-image}). We call the cropped
out images \emph{single plant images }and the original images \emph{multiple
plant images}. Further, a copy of the original multiple plant image
is created and annotated with the boundaries of the cropped out single
plant image. The four selections that can be made in the second section
of the interface relate to these three image types and the respective
metadata, which is given in a JSON file. Selecting several filetypes
will result in all checked filetypes to be returned from the server.
Finally, the second section contains a drop-down list that contains
pre-compiled datasets. These datasets are permanently stored on the
server-backend (in contrast to user-defined datasets that are being
pulled from the object storage) and thus can be delivered immediately
by the server.

Once the user has defined a dataset through the first two sections,
they can query the server in three different ways through the third
section of the interface. 
\begin{itemize}
\item First, the user can check how many files fulfill the specified filter,
without committing to any downloads, by using the ``Check Query''
button. The information returned by the server is displayed in the
TC as the number of files and in how many parts these files would
be downloaded. 
\item By using the ``Get Sample'' button the TC will download a small
randomly selected sample per selected filetype. For example, if ``Single
Plant Images'' and ``Multiple Plant Images'' is selected the TC
will download 20 random images of each filetype. 
\item Finally, the user can download all files resulting from a query by
using the ``Download'' button. This will download all files that
fit the specified filters and filetypes. 
\end{itemize}
Three additional buttons are available: 
\begin{itemize}
\item ``Set Credentials'': This opens a dialog in which the user can enter
or change their user credentials consisting of username and password.
The credentials are saved locally and are reused at future launches
of the TC. 
\item ``Set Sample Path'': This allows the user to select in which folder
samples should be saved to. 
\item ``Set Download Path'': This allows the user to select in which folder
downloads through the ``Download''-button will be saved to. 
\end{itemize}
The functionality of the TC when using the ``Fielddata''-tab is
similar to the one under the ``EAGL-I Data''-tab. Future datasets
that will be available through the TC client will be accessed through
additional tabs, one per dataset, each containing a similar interface. 

\section{Performance\label{sec:Performance} }

The TC's performance depends on several factors that lie outside of
its control:
\begin{itemize}
\item The end-systems download-speed and write-speed to drive
\item The current load on Compute Canada's internal network
\item The number of clients connected and using the server back-end
\end{itemize}
To investigate the impact of parallel usage, we tested the system
with up to 6 users downloading files at the same time. Only 2 users
(user 2 and user 3) shared the same access network. In this test we
asked the users to download a predefined dataset of 6193 (542 MB)
and measured the time until the data was available on their system.
This was compared to the time it took user 1 to download the same
files, while no other user is active, which took 130 seconds. We saw
that the impact of parallel usage was measurable, yet small, compared
to the impact the end-to-end bandwidth between the client's host and
the server has (see Table \ref{tab:Performance-Test-of}). For example,
users 2, 3, and 6 encountered a similar download time independent
of whether they downloaded pre-compiled or not pre-compiled datasets.
This tells us that the bottleneck in their download speed was not
the process of fetching data from the object-storage. Indeed, user-defined
(not pre-compiled) datasets impose additional workload on the server
that consists of retrieving the data from the object storage and tar-balling
the files before sending them to the TC (see the next Section for
details). We also saw that downloading pre-compiled datasets is significantly
faster if the end-user has enough download bandwidth. 

In our testing we further noticed that the performance of the TC and
the server-backend fluctuates even if the available bandwidth between
client and server remains stable and no other clients are active.
We have observed a decrease of download speed up to a factor of 2
in rare instances. When investigating these temporary drops in performances
we noticed that the transfer-speed between object storage and server-backend
lead to these longer delays. We thus assume that the overall load
on Compute Canada's internal network can impact the system's overall
performance. 

\begin{table}

\begin{centering}
\caption{\label{tab:Performance-Test-of}}
\par\end{centering}
\begin{centering}
\begin{tabular}{|c|c|c|>{\centering}p{2cm}|}
\hline 
 & Download Time & Download Speed & Download Time\break (precompiled)\tabularnewline
\hline 
\hline 
User 1 & 316 seconds & 1.7 MB/s & 36 seconds\tabularnewline
\hline 
User 2 & 957 seconds & 0.56 MB/s & 840 seconds\tabularnewline
\hline 
User 3 & 1,006 seconds & 0.53 MB/s & 847 seconds\tabularnewline
\hline 
User 4 & 1096 seconds & 0.49 MB/s & 49 seconds\tabularnewline
\hline 
User 5 & 300 seconds & 1.8 MB/s & 58 seconds\tabularnewline
\hline 
User 6 & 972 seconds & 0.55 MB/s & 965 seconds\tabularnewline
\hline 
\end{tabular}
\par\end{centering}
\smallskip{}

\begin{centering}
Performance Test of 6 users downloading 6193 files (542 MB) in parallel
\par\end{centering}
\end{table}

We also performed a long running speed-test for downloading 95K single
plant images (9.7 GB) for a single active user, where the data was
not precompiled. We repeated that download at several different days
to rule out fluctuations stemming from Compute Canada's internal traffic.
The results had been consistent within 1 minute. Similarly we downloaded
20K multiple plant images (34.3 GB) and 15K field data images (105
GB) to compare the download speeds with respect to different filetypes
(single plant images are smaller files). The results can be found
in Table \ref{tab:Long_run_test}. We can observe that the end-to-end
download speed for single plant images lies far below the download
speed for the other two categories. The reason for that is the overhead
of copying many indiviual small files from the object storage to the
server-backend within Compute Canada's network. For each file we can
observe a significant overhead spent on initializing the transfer,
resulting in poorer performance. 

\begin{table}
\begin{centering}
\caption{\label{tab:Long_run_test}}
\par\end{centering}
\begin{centering}
\begin{tabular}{|c|c|c|}
\hline 
 & Download Time & Download Speed\tabularnewline
\hline 
\hline 
Single plant images & 80 minutes & 2 MB/s\tabularnewline
\hline 
Multiple plant images & 34 minutes & 18 MB/s\tabularnewline
\hline 
Field data images & 95 minutes & 19 MB/s\tabularnewline
\hline 
\end{tabular}
\par\end{centering}
\smallskip{}

\begin{centering}
Performance test for data from different categories.
\par\end{centering}
\end{table}

\section{Architecture\label{sec:Architecture} }

In this section we give a high-level description of the client and
server architecture, starting with the client. 

\subsection{Client Architecture}

We focus the discussion on how downloading a user-defined dataset
is implemented on the client side. Downloading samples instead or
querying the database without starting a download at all are just
variations and simplifications of a full download. 

The main functionality is defined in $\mathtt{simple\_client.py}$,
which provides two key methods: $\mathtt{send\_req}$ and $\mathtt{get\_files}$.
The former is a wrapper to send HTTPS POST- and GET-requests to the
server's API. It is used to communicate requests to the database or
request a file to be downloaded. In the former case the return of
this method is a JSON-formatted object containing the response of
the server. When the user wishes to download a dataset $\mathtt{send\_req}$
is used to inform the server about the query parameters. In response
the client obtains a unique randomized job-id together with the number
of parts the download will consist of. Then the TC main-loop starts
the $\mathtt{get\_files}$ method as a subprocess for this job-number
and the list of parts.

The $\mathtt{get\_files}$ method repeatedly asks the server if a
specific part of a specific job-id is ready to download. To not overload
the server with these queries a dynamic backoff period is used, which
increases between successive requests. This backoff period resets
after a request yields a positive answer from the server. If the server
answers that the part is ready for download, $\mathtt{get\_files}$
continues by a direct request to the tar-file via $\mathtt{send\_req}$.
The such received data is then unpacked into the download folder on
the client side (specified by the user (see Section \ref{sec:Operation}).
If the server repeatedly answers that the data is not ready, the $\mathtt{get\_files}$
method will eventually reach a maximal number of tries and abandon
the download of that part. Once a part is downloaded (or the maximal
number of tries exceeded) the main-loop ends that subprocess and moves
on to download the next part in a new subprocess. This is repeated
until all parts are processed.

Since the main functionality of the client is contained within $\mathtt{simple\_client.py}$
an operation outside the provided graphical user interface is possible.
Indeed, the graphical user interface is just a convenient way to construct
well-formed HTTPS requests to the server. It is thus possible to skip
its usage entirely for the purpose of accessing the server-side API
directly within custom code. This allows, for example, to fetch data
from the server on-demand from within other methods, such as the iterative
training of machine learning models. 

\subsection{Server Architecture}

The server is hosted on Compute Canada on an end-system separate from
the data object storage. For communication with the clients the server
runs a stack of Flask, MongoDB, and Celery. Flask provides an API
over HTTPS, which is also used to deliver the data to the clients.
Besides requests that are related to user-management the server handles
two types of requests: database queries and download requests. 

Database queries from the client are a collection of query parameters
which are sanitized and transformed into MongoDB-queries server-side.
The such formed MongoDB-query is used to obtain the list of files
the user wishes to download in the future. The server also creates
a job-id and allocates resources for the upcoming download process.
A summary of the file-list is sent back to the client immediately,
together with the job-id. Then, using Celery, the server starts a
subprocess that downloads the first part of the requested data from
the object storage. Once the first part is fully fetched from the
object storage those files are tar-balled and are now available for
download by the client. Without waiting for the client, the server
immediately starts the download of the second part following the same
process. However, the third part will only be fetched from the object
storage after the first part has been downloaded by the client and
the tar-archive has been deleted at the server-instance. The server
continues this double-buffered approach, downloading new parts from
the object-storage while serving the previously fetched part to the
client (see Figure \ref{fig:Architecture}). This architecture has
two advantages: First, it serves the first parts after a short time
to the client thus being a responsive design. Second, we do not have
to download the requested dataset in its entirety to the server-instance.
In fact, this would be impossible for datasets that exceed the server-instance's
storage (the object storage can host hundreds of TB of data, whereas
the server instance has a limited disk space of 5 TB). 

\section{Future Work\label{sec:Conclusion-and-Future}}

We presented the TerraByte Client a tool for downloading user-defined
datasets from the TerraByte data-server. We continue our goal to provide
labelled plant-data to researchers and industry with the purpose of
progressing digital agriculture. To that end, we will extend the client
as well as the data-portal along several dimensions:
\begin{itemize}
\item More variety in field-data: Future iterations of the data-portal will
contain drone data of field images, including test-plots of research
facilities and fields of local farmers. Further, we are developing
a data-rover that will collect image data in test-plots. More plant-variety
in field-data is planned as well. 
\item Beyond RGB-data: We have several systems that go beyond simple RGB-imaging
that are being prepared to ingest data on a large into our database.
Amongst these systems are a 3D-scanner for phenotyping purposes, a
hyperspectral camera, and a photogrammetry station. 
\item Additional imagers and plant variety: Next to continuing the operation
of EAGL-I, we are setting up additional imaging systems that will
image a wider variety of crops and weeds inside our laboratories. 
\end{itemize}
We will further work on improving the overall performance of the client
and the server. Specifically, we will increase the end-to-end download
speed for user-defined datasets. One options is to have datasets prepared
and saved server-side, so that they can be downloaded later at a faster
speed. For this purpose, each user will have a limited allocation
of disk-space on the server, they can still continue to use the client
in the way outlined above as well without space-limitations. Another
options is to create archives on the object storage each containing
a subset of the data, such that if one file inside a subset is queried
it is very likely that other files from that archive are also queried
(e.g., one specific day of imaging). Since the data querried usually
is highly correlated along such a clustering, transferring an archive
from the object storage to the server-backend can yield many relevant
files while only paying the transfer overhead once. Indeed, we observed
in tests a tranfer speed of more than 150 Mb/s from object storage
to the server instance. This would shift the bottleneck of end-to-end
transfer to the connectin between the TC and the server instance (for
which we observed upload speeds of 20 Mb/s). Further improvements
can be achieved by implementing a cache on the level of downloaded
archives and a cache on the level of extracted (or copied) files to
speed-up follow-up queries. 

\bibliographystyle{unsrt}
\bibliography{refs/references}

\end{document}